\documentclass{article}

\usepackage[final]{corl_2020} 
\usepackage{graphicx}
\graphicspath{{cna/}}
\usepackage{amsmath}
\usepackage{amsfonts}
\usepackage{booktabs}
\usepackage{hyperref}
\usepackage{pgfplots}
\usepackage{xfrac}
\pgfplotsset{width=7cm,compat=1.8}
\usetikzlibrary{spy}

\title{Belief-Grounded Networks for Accelerated\\Robot Learning under Partial Observability}

\author{
  Hai Nguyen\footnotemark[1],\ \ Brett Daley\footnotemark[1],\ \ Xinchao Song,\ \ Christopher Amato\footnotemark[2],\ \ Robert Platt\footnotemark[2]\\
  Khoury College of Computer Sciences\\
  Northeastern University, Boston, MA 02115\\
  \texttt{\{nguyen.hai1, b.daley, song.xin, c.amato, r.platt\}@northeastern.edu} \\
}
\DeclareMathOperator*{\E}{\mathbb{E}}
\usepackage{algorithm}
\usepackage{algpseudocode}
\usepackage{mathtools}

\newcommand{\tightsection}[1]{
    \vspace{-0.075in}
    \section{#1}
    \vspace{-0.1in}
}
\newcommand{\tightsubsection}[1]{
    \vspace{-0.0375in}
    \subsection{#1}
    \vspace{-0.05in}
}
\newcommand{\tightparagraph}[1]{
    \vspace{-0.12in}
    \paragraph{#1}
}
\newcommand{\numcircled}[1]{\raisebox{.5pt}{\textcircled{\raisebox{-.9pt} {#1}}}}

\newcommand{\blfootnote}[1]{%
  \begingroup
  \renewcommand\thefootnote{}\footnote{#1}%
  \addtocounter{footnote}{-1}%
  \endgroup
}

\begin{document}

\begin{NoHyper}  
    \maketitle
\end{NoHyper}
\setcounter{footnote}{0}  


\begin{abstract}
    Many important robotics problems are partially observable where a single visual or force-feedback measurement is insufficient to reconstruct the state.
    Standard approaches involve learning a policy over beliefs or observation-action histories.
    However, both of these have drawbacks;
    it is expensive to track the belief online, and it is hard to learn policies directly over histories.
    We propose a method for policy learning under partial observability called the Belief-Grounded Network (BGN) in which an auxiliary belief-reconstruction loss incentivizes a neural network to concisely summarize its input history.
    Since the resulting policy is a function of the history rather than the belief, it can be executed easily at runtime.
    We compare BGN against several baselines on classic benchmark tasks as well as three novel robotic force-feedback tasks.
    BGN outperforms all other tested methods and its learned policies work well when transferred onto a physical robot.
    
    \blfootnote{\hspace{-1.8ex} * \hspace{-1ex} Equal contribution.}
    \blfootnote{\hspace{-1.55ex} \textsuperscript{\textdagger} \hspace{-1ex} Equal advising.}
\end{abstract}



\section{Introduction}

Touch is an important sensory modality in robotics because it is often more precise than vision and is not subject to occlusions.
However, because a sequence of touch observations is typically needed to make sense of the environment, using touch requires solving a challenging Partially Observable Markov Decision Process (POMDP).
Offline POMDP planning methods often use a \textit{belief} over states instead, which is known to be a sufficient statistic for the observation history (\textit{e.g.}, \citep{Kurniawati_2008, silver2010monte, smith2012heuristic, somani2013despot}).
Unfortunately, explicitly tracking beliefs online can be computationally expensive;
while we may be willing to track beliefs during training, we do not want to do it online.
If we could instead learn a policy over histories of observations and actions, we would obviate the need for belief tracking at runtime.
In fact, this is exactly what Deep Recurrent Q-Network (DRQN)~\cite{hausknecht2015deep} does.
The Q-function is expressed as a function of the observation history rather than the belief, but it often learns slowly because it is difficult to extract high-quality features from the history using only the reinforcement learning loss.
Ideally, we would leverage the ability to track beliefs at training time in order to learn a history-based policy faster. 

This paper proposes an approach to achieve exactly this.
We introduce a new model called the Belief-Grounded Network (BGN) where we add a belief-reconstruction loss to a deep reinforcement learning agent during simulated training.
Thus, the neural network is incentivized to generate features that summarize the history, and the agent autonomously learns to balance information gathering with task-specific objectives in an end-to-end manner.
After training, the policy can be deployed without belief access since only the history is needed for the network's forward pass.
We test BGN in four classic POMDPs and three novel force-feedback manipulation tasks, where it outperforms a number of strong baselines.
Not only does BGN outperform Asymmetric Actor-Critic \cite{pinto2017asymmetric}, but it even outperforms an agent that learns directly from the true belief representation.
We demonstrate\footnote{
    Code and videos are available at \url{https://sites.google.com/view/bgn-pomdp/home}
}
that the policies learned in simulation by BGN work well when transferred to a physical robot without any adjustments.

\tightsection{Background}

When an agent cannot fully ascertain the underlying state of its environment, the problem can be modeled as a POMDP~\cite{kaelbling1998planning}.
Formally, a POMDP can be specified by the tuple
$(\mathcal{S}, \mathcal{A}, \mathcal{T}, \mathcal{R}, \Omega, \mathcal{O})$.
At time $t$, the environment exists in some state $s_t \in \mathcal{S}$ that can be manipulated by the agent's chosen action $a_t \in \mathcal{A}$.
The resultant state $s_{t+1} \in \mathcal{S}$ is sampled with probability
$\smash{ T(s_t,a_t,s_{t+1}) }$
and the agent receives a reward
$\smash{ r_t = \mathcal{R}(s_t,a_t,s_{t+1}) }$.
This process repeats until the episode ends.
The goal of the agent is to take actions according to a policy $\pi$ that maximize its expected discounted return~\cite{sutton2018reinforcement} defined as
$\E_\pi[\sum_{t=0}^\infty \gamma^t r_t]$
for some discount factor $\gamma \in [0,1]$.

It is not possible for the agent to directly observe the state in a POMDP.
Instead, the agent receives an observation $o_t \in \Omega$ which is indirectly related to $s_{t+1}$ via the observation function
$\mathcal{O}(s_{t+1},a_t,o_t) = P(o_t \mid s_{t+1},a_t)$.
In general, this implies that the agent must take the entire history of observations and actions
${h_t = ((a_0, o_0), (a_1, o_1), \ldots, (a_{t-1}, o_{t-1}))}$
into account to make optimal decisions.

Because the set of possible histories grows exponentially with time, exact history-to-action mappings are typically intractable.
A history summary $\smash{\hat{h}_t}$ can be substituted, but this may preclude optimal behavior.
The alternative is to maintain a belief $b \in \Delta(\mathcal{S})$ over possible states,
where $b(s)$ denotes the probability that the environment's true state is
$s \in \mathcal{S}$.
The belief is a sufficient statistic of the complete history, allowing for the construction of optimal policies~\cite{kaelbling1998planning}.
It is straightforward to compute the next belief $b_{t+1}$ from the current belief $b_t$:
\begin{equation}
    \label{eq:belief_update}
    b_{t+1}(s_{t+1}) \gets \frac{\mathcal{O}(s_{t+1}, a_t, o_t) \sum_{s \in \mathcal{S}} T(s, a_t, s_{t+1}) b_t(s)}{P(o_t \mid a_t, b_t)}
\end{equation}
The Markov property of the belief suggests that it should facilitate better policies than would an incomplete history summary.
In Section \ref{sec:bgn}, we introduce a general method for obtaining these training benefits in simulation while avoiding the belief update in Equation (\ref{eq:belief_update}) during execution.

\tightsubsection{Advantage Actor-Critic (A2C)}
\label{sec:a2c}
Advantage Actor-Critic (A2C) \cite{dhariwal2017openai} is a synchronous version of A3C \citep{mnih2016asynchronous}.
Multiple agents alternate between sampling short sequences of experience and updating a shared parametric function, with each agent operating in a distinct environment instance to help decorrelate the training experience.
The parameters can be factored into two subsets $\theta$ and $\phi$ according to their functionalities;
the \textit{actor} samples an action $a_t$ with probability determined by the policy $\pi(a_t \mid s_t; \theta)$, while the \textit{critic} estimates the discounted return $V(s_t; \phi)$ expected by executing $\pi$ from state $s_t$.
Using backpropagation \cite{rumelhart1986learning}, the parameters are jointly optimized by reducing a three-term loss averaged over each timestep $t$ of the trajectory.
The loss gradient is defined as the following:\footnote{
    When multiple agents are present, their respective gradients are simply summed together at each timestep.
}
\begin{equation}
    \label{eq:a2c_loss}
    \nabla \mathcal{L}_t
        = \nabla_\theta \mathcal{L}_t^{(\text{actor})}
        + \nabla_\phi \mathcal{L}_t^{(\text{critic})}
        + \nabla_\theta \mathcal{L}_t^{(\text{entropy})}
\end{equation}
Here, $\smash{ \mathcal{L}_t^{(\text{actor})} }$ increases the log-likelihood of actions that are better than expected;
$\smash{ \mathcal{L}_t^{(\text{critic})} }$ improves the critic by reducing the squared error between its prediction and the return estimate $R_t$ computed from the current experience sequence;
and $\smash{ \mathcal{L}_t^{(\text{entropy})} }$ discourages premature convergence to a deterministic policy.
Mathematically, these losses are defined in the following manner:
\begin{equation*}
    \smash{
        \mathcal{L}_t^{(\text{actor)}} = - \log \pi(a_t \mid s_t; \theta) (R_t - V(s_t; \phi))
        \quad\quad\quad
        \mathcal{L}_t^{(\text{critic})} = \beta_c (R_t - V(s_t; \phi))^2
    }
\end{equation*}
\vspace{-0.1in}
\begin{equation*}
    \smash{
        \mathcal{L}_t^{(\text{entropy})} = -\beta_e H(\pi(\cdot \mid s_t; \theta))
    }
\end{equation*}
The coefficients $\beta_c, \beta_e > 0$ control the relative magnitudes of these terms, and $H$ represents the information entropy.
Because the state $s_t$ is generally obfuscated in our problem setting, we note that these losses can be trivially adapted to accept a history summary $\smash{\hat{h}_t}$ or the belief $b_t$ in place of $s_t$.
We will do this frequently throughout our work.

\tightsection{Belief-Grounded Networks (BGNs)}
\label{sec:bgn}

The belief perfectly summarizes the observation-action history in the sense that no additional information is needed to behave optimally in a POMDP.
We therefore expect that a policy based on the belief would perform better than one that substitutes some alternative (incomplete) summary.
Nevertheless, there are practical concerns that preclude the use of beliefs in robot applications.
The first is that tracking the belief is cumbersome for online execution.
Policies that rely on the belief must constantly invoke Equation (\ref{eq:belief_update}) just to select actions, which is expensive.
The second concern is that the belief is an \textit{unstructured} representation in the sense that it is unclear how to induce a deep learning model to reasonably generalize to novel beliefs.
In practice, it may actually prove easier to learn from partial observations when they can be encoded using appropriate neural network structures---for example, images with convolutional layers.

These challenges lead us to seek a general technique that can benefit from beliefs while training, but does not need beliefs for action selection.
Our proposed solution is to train an agent to \textit{reconstruct} the belief from incomplete observations in conjunction with the standard reinforcement learning problem.
We call our method the Belief-Grounded Network (BGN) because the learned model is guided by ground-truth beliefs in simulation.
Let $\smash{f(\hat{h})}$ denote the \textit{features} extracted from an arbitrary history summary $\smash{\hat{h}}$ by a nonlinear parametric model $f$.
Deep reinforcement learning agents typically use a linear combination of these features to parameterize a stochastic policy.
By minimizing a loss measure $\mathcal{L}$ with respect to the parameters of $f$, an agent can improve its performance at a given control task.
We propose an augmented loss $\mathcal{L}^+$ that simultaneously trains the agent to summarize its input history:
\begin{equation}
    \label{eq:belief_loss}
    \smash{ \mathcal{L}^+ = \mathcal{L} + H(b, \hat{b}) }
\end{equation}
The additional term $\smash{H(b, \hat{b})}$ denotes the cross entropy of a reconstructed belief $\smash{\hat{b}}$ relative to the true belief $b$.
The former is predicted from the features $\smash{f(\hat{h})}$ and the latter is provided to the agent by the simulator.
Thus, Equation~(\ref{eq:belief_loss}) has an appealing mathematical significance.
Because the entropy of the true belief $H(b)$ is constant with respect to the parameters of $f$, minimizing the cross entropy is equivalent to minimizing the Kullback-Leibler divergence
$\smash{ D_{\text{KL}}(b \; \| \; \hat{b}) = -H(b) + H(b, \hat{b}) }$.
Hence, this auxiliary task generates a maximum-likelihood estimate for the true belief $b$ conditioned on a particular history summary $h$.
If this estimate is accurate, then the features $\smash{ f(\hat{h}) }$ should also be a sufficient statistic for the history \cite{kaelbling1998planning}.
We expect this learned representation to be better than directly using the belief $b$ because the former can be arbitrarily adapted for the task at hand.

\begin{figure}[t]
    \centerline{\includegraphics[scale=0.7]{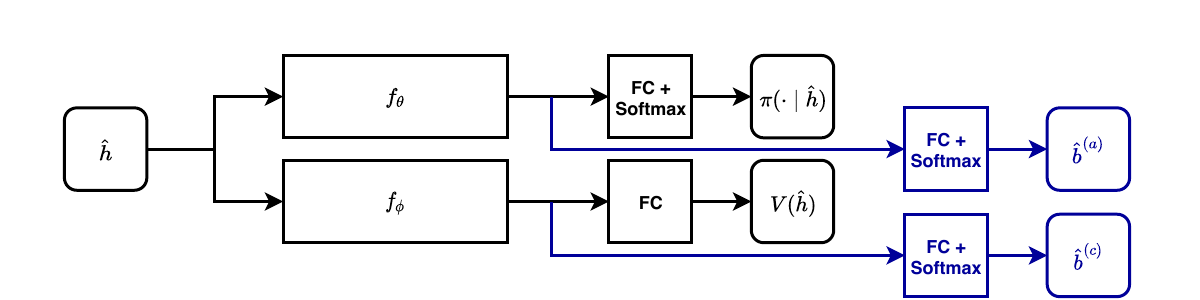}}
    \caption{
        A schematic of our Belief-Grounded Network (BGN) combined with A2C.
        The belief is reconstructed from the neural networks' features to aid learning under partial observability.
    }
    \label{fig:belief_grounded_network}
\end{figure}

Let us exemplify the BGN by combining it with A2C.
Assuming that the actor and critic do not share parameters, we require two BGNs for the separate feature extractors:
$\smash{f_\theta(\hat{h})}$ and $\smash{f_\phi(\hat{h})}$, respectively.
This architecture is depicted in Figure \ref{fig:belief_grounded_network}.
Black components represent a standard dual-network A2C agent, while blue components indicate the additional network heads for reconstructing the belief.
``FC'' stands for a fully-connected network (\textit{i.e.}, a linear combination).
We assume that observations and actions are discrete, and therefore use softmax activation functions for the policy distribution (``Distr.'') and the reconstructed beliefs $\smash{b_t^{(a)}}$ and $\smash{b_t^{(c)}}$.
In continuous environments, the softmax can be substituted for other families of distributions.
Training proceeds as described in Section \ref{sec:a2c}, but the gradient is reformulated by combining Equations (\ref{eq:a2c_loss}) and (\ref{eq:belief_loss}):
\begin{equation}
    \label{eq:a2c_belief_loss}
    \nabla \mathcal{L}_t^+ = \nabla \mathcal{L}_t
        + \nabla_\theta H(b_t, \hat{b}_t^{(a)})
        + \nabla_\phi H(b_t, \hat{b}_t^{(c)})
\end{equation}
The BGN can be used in conjunction with a variety of learning algorithms.
While our later experiments will focus on A2C, the BGN could be other combined with other actor-critic methods like DDPG \cite{lillicrap2015continuous} and PPO \cite{schulman2017proximal} or value-based methods like DQN \cite{mnih2015human} and DRQN \cite{hausknecht2015deep}.
In all cases, the resulting policies operate solely with histories, making them appropriate for robot applications where tracking the belief is undesirable or infeasible.

\tightsection{Related Work}

Using privileged information during simulation to efficiently generate policies for real-world, partially observable environments is a recent trend in robotics.
Other works have principally focused on exploiting underlying state information to aid with representation learning, particularly in visuomotor control tasks like image-based manipulation \cite{pinto2017asymmetric, wilson2020learning} and urban driving \cite{chen2020learning}.
While this may help the agent in these limited cases, it is easy to see how this approach would fail when the state representation is less interpretable.
This is precisely the case for Atari 2600 games, where augmenting the visual input of DQN with the fully observable RAM state has been shown to be ineffective \cite{sygnowski2016learning}.
In contrast, the belief's consistent probabilistic meaning across all possible POMDPs makes our information-theoretic loss agnostic to the precise interpretation of the underlying state.

A number of works have explored the use of auxiliary tasks to enhance learning, but ours is the first to use beliefs for this purpose.
Other approaches have aimed to incentivize exploration in sparse-reward environments \cite{pathak2017curiosity} or develop better feature representations for more sample-efficient learning \cite{jaderberg2016reinforcement, mirowski2016learning, shelhamer2016loss, srinivas2020curl}.
The BGN falls into the latter category.
We also note that our auxiliary task is analogous to the theory of predictive state representations (\textit{e.g.}, \cite{littman2002predictive, singh2004predictive, baisero2020learning}), but we reconstruct the belief instead of rewards and observations.
The advantage of our method is that the belief is a sufficient statistic for the history, whereas a sequence of reward-observation pairs is generally not.
Nevertheless, all of these other methods are orthogonal to BGN and could be combined for potentially better performance in difficult POMDP environments.

Finally, we highlight some recent works that apply the POMDP framework to similar manipulation problems.
Unlike our end-to-end learning approach, these methods utilize expensive, hard-coded techniques to facilitate solutions:
\textit{e.g.}, POMDP solvers \cite{vien2015pomdp, garg2019learning}, belief tracking during execution \cite{wirnshofer2020controlling}, Bayesian updates \cite{sung2017learning}, or traditional controllers \cite{beltran2020learning}.
In contrast, by specifying only an augmented loss during training, the BGN is granted the flexibility to generate a representation that is best suited to the task at hand.

\tightsection{Experiments}

Training a deep reinforcement learning agent with BGN should help it learn faster under partial observability by encouraging extracted features to summarize the observation-action history.
To test this, we compare the BGN against various related methods when combined with an A2C agent (see Appendix \ref{app:training-details} for hyperparameters).

We adopt a shorthand notation for agent names in the form Ax-Cy, where x and y represent the respective inputs for the \textbf{A}ctor and the \textbf{C}ritic.
The possible input types are \underline{h}istory summary, \underline{b}elief, and \underline{s}tate.
For example, \textbf{Ah-Ch} denotes a standard A2C agent that uses history summaries.\footnote{
    We generate history summaries with Gated Recurrent Units \cite{cho2014properties};
    see Appendix \ref{app:agents} for architectures.
}
By this convention, we refer to our new method from Section \ref{sec:bgn} as \textbf{Ah-Ch + BGN}.

The other baselines are as follows.
\textbf{Ah-Cs}:\
Asymmetric Actor-Critic \cite{pinto2017asymmetric} where the critic uses ground-truth states provided by the simulator, while the actor uses observation-action histories with a recurrent network.
\textbf{Ah-Cb}:\
An asymmetric agent similar to Ah-Cs, but the critic uses beliefs instead of states.
Theoretically, this should reduce bias in the case where two different observation-action histories arrive at the same state.
\textbf{Ab-Cb}:\
Both the actor and the critic accept the belief as input;
because the policy depends on the belief, this cannot be executed on a robot easily.
\textbf{SARSOP}:\
One of the leading offline POMDP planning methods that uses beliefs \citep{Kurniawati_2008}.
This offers an approximate idea of the best attainable performance for each environment.
\textbf{Random}:\
An untrained agent that selects actions from a uniform distribution over $\mathcal{A}$.

\tightsubsection{Classic POMDPs}
\label{subsec:classic}
We begin by testing our method in four classic POMDPs that are challenging due to their stochasticity and partial observability.
We provide brief descriptions of each environment below, with additional details in Appendix \ref{app:pomdp-details}.

\tightparagraph{Hallway and Hallway-2}
An agent must reach a fixed goal cell starting from a random location in a gridworld, where the observation and transition models are extremely noisy \cite{littman1995learning}.
States consist of the agent's position and orientation, and the goal location.
Observations indicate the presence of a wall in each of the four directions.

\paragraph{RockSample[4,4] and RockSample[5,5]}
RockSample$[n,k]$ simulates a rover exploring an ${n \times n}$ gridworld with $k$ rocks at predefined locations \citep{smith2012heuristic}.
The rover starts at a fixed location and must find good rocks to get rewards.
The state includes its current coordinate and $k$ binary features that indicate which rocks are good.
The agent can check the quality of a rock at its current location or from a distance using a sensor that returns a noisy binary observation.

\begin{figure}[t]
    \centering
    \includegraphics[width=0.75\linewidth]{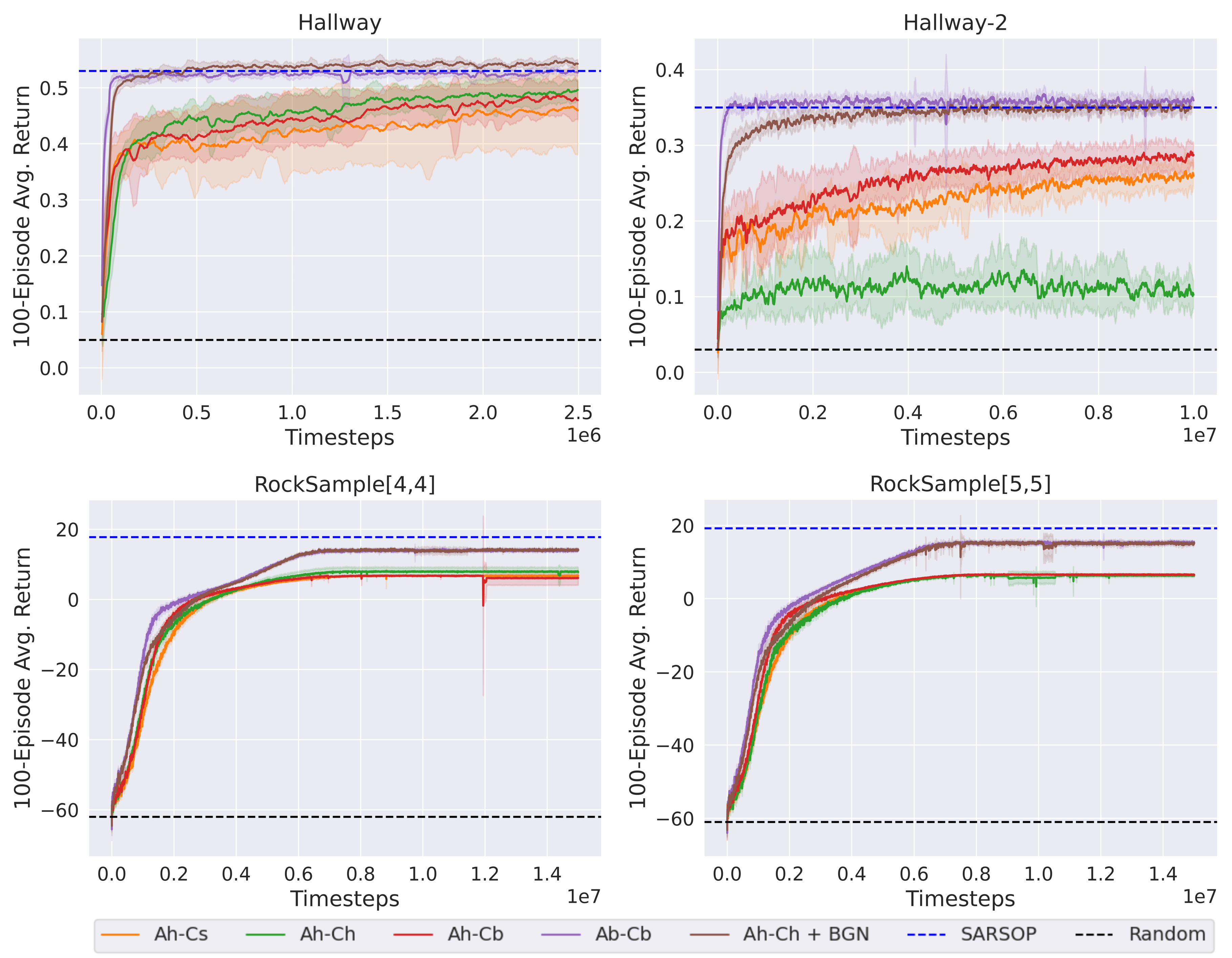}
    \caption{
        Training performance for each classic POMDP domain.
    }
    \label{fig:result-1}
\end{figure}

\tightparagraph{Results}
We plot the mean learning curves (100-episode moving average) of all methods over 10 random seeds with standard deviations shaded in Figure \ref{fig:result-1}.
Readers can refer to Appendix \ref{app:raw-score} for the raw and normalized returns of the final policies.
Across all domains, we observe that the methods bifurcate into two groups according to their performances.
In the better-performing group, Ah-Ch + BGN and Ab-Cb perform similarly in each environment, matching SARSOP's performance in the Hallway domains and coming close to it in the RockSample domains.
In the other group, the performance order is inconsistent;
while Ah-Cb's use of the belief helps it outperform Ah-Cs in Hallway and Hallway-2, it appears to not have a significant effect in the RockSample domains.
Additionally, despite not having access to any privileged information, Ah-Ch is able to match Ah-Cb and Ah-Cs on three of the four environments.
This suggests that asymmetric architectures may have limited benefits under extreme levels of partial observability.

\tightsubsection{Force-Feedback Robotic Domains}
\label{subsec:tactile}

\begin{figure}[t]
    \centering
    \includegraphics[scale=0.53]{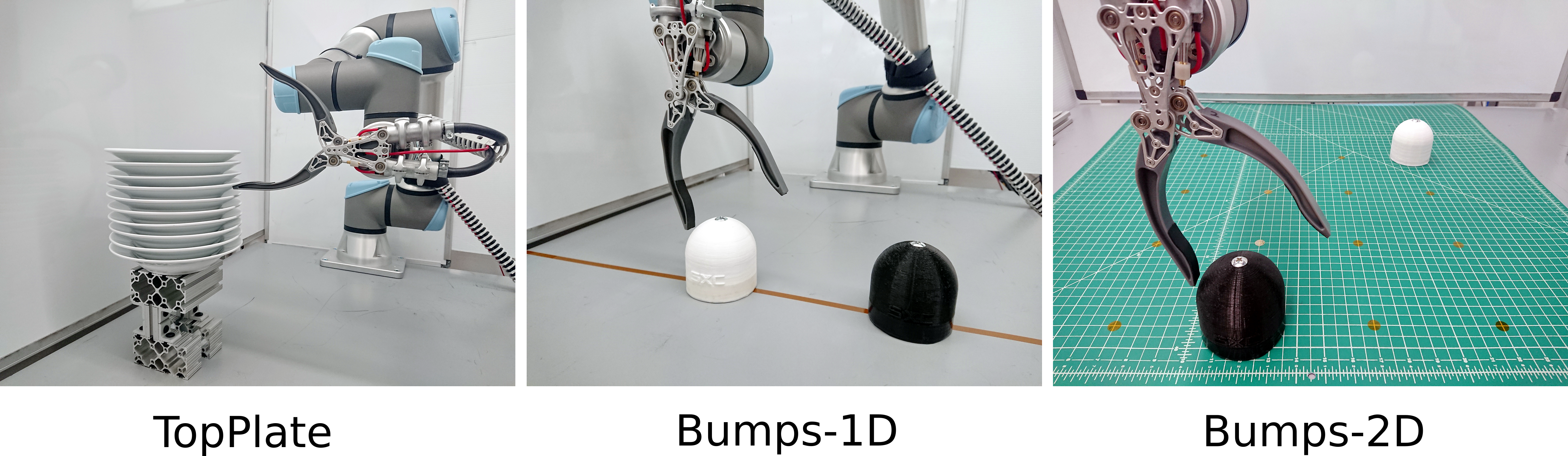}
    \caption{
        Our three force-feedback robotic domains.
        The agent controls a finger with an adjustable stiffness to first localize and then manipulate an object of interest.
    }
    \label{fig:robot-setup}
\end{figure}

We implement three novel force-feedback robotic tasks in simulation (MuJoCo \cite{todorov2012mujoco}) and on a physical robot (Figure \ref{fig:robot-setup}).
The fingers of the robot have controllable compliance: they can be compliant or stiff as specified by the agent action command.
In compliant mode, the fingers glide over the objects while in stiff mode they can grasp or push.
These tasks are classified as Mixed Observability MDPs (MOMDPs) \citep{ong2009pomdps} (see Appendix \ref{app:momdp}) because the agent can observe the coordinate and angle of its finger perfectly but it does not observe the positions of the objects.
All reward functions are sparse;
agents receive a reward of $+1$ only upon successful task completion and $0$ otherwise.
We discretize the state and observation spaces to compute beliefs efficiently.
We briefly described each task below with additional details provided in Appendix \ref{app:momdp-details}.

\tightparagraph{TopPlate}
(Figure \ref{fig:robot-setup}, left).
The agent must locate and grasp the top plate from a stack of height $k \sim U([1,10])$ where the height of the stack is initially unknown to the agent.
It locates the top plate using only finger position observations when the finger is in compliant mode.
The agent's finger is positioned such that it touches the plates as it moves up and down in compliant mode.
When ready, the agent can execute a command to grasp the plate that is adjacent to it.
The episode terminates when a grasp action is performed.

\tightparagraph{TwoBumps-1D}
(Figure \ref{fig:robot-setup}, middle).
Two movable bumps rest on a table, with the robot's finger moving along a horizontal line above them.
The initial position of the robotic finger as well as the two objects are randomized uniformly such that the left-right order between the bumps is unchanged.
The agent's goal is to push the rightmost bump to the right without disturbing the left bump.
There are four action combinations:
move left or right, each with a compliant or stiff finger.
This task is challenging because the agent does not know initially which bump is rightmost---it must touch both bumps to determine this.
Because the agent's motion is constrained one-dimensionally, it is not possible to miss the bumps.
The robot must relax the stiffness of its finger when passing by a bump to avoid pushing the wrong one.
The episode ends as soon as either bump moves.

\tightparagraph{TwoBumps-2D}
(Figure \ref{fig:robot-setup}, right).
Two bumps of different sizes are randomly positioned on a ${4 \times 4}$ grid.
The robotic finger is constrained to move in a plane above the bumps. The finger can be moved in any of the four directions or perform a grasp.
The agent must make contact with each bump at least once and then grasp the larger bump to complete this task successfully, inferring the bumps' relative sizes from the angular displacement of the finger.
The episode ends after a grasp is executed.

\begin{figure}
    \centering
    \includegraphics[scale=0.6]{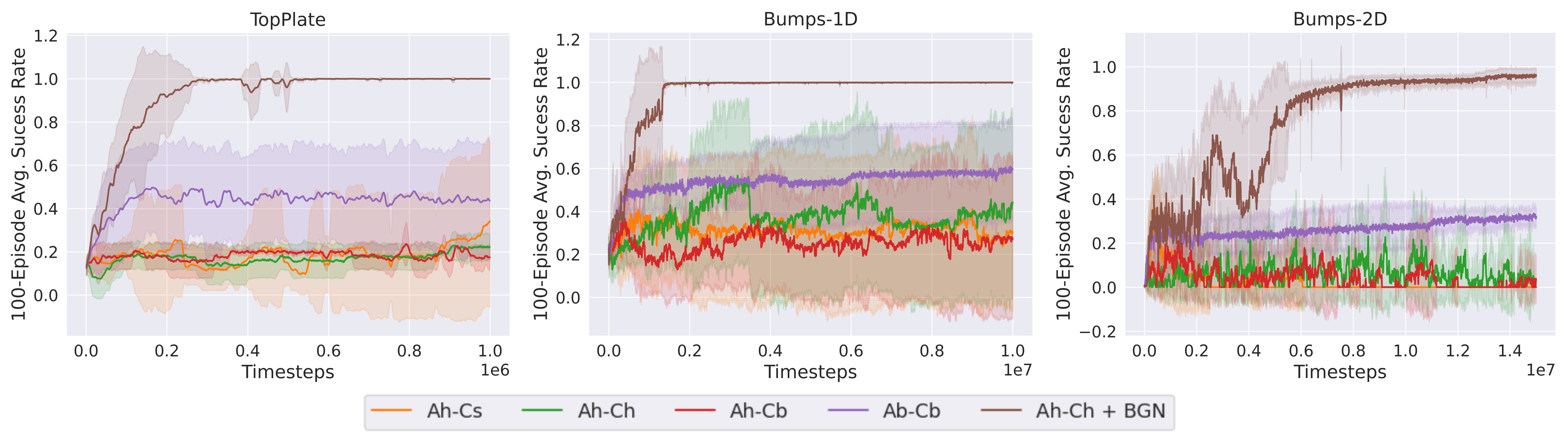}
    \caption{The average success rates of all methods in three robotic tasks.}
    \label{fig:momdp-results}
\end{figure}

\tightparagraph{Results}
We plot the success rates (100-episode moving average) of the methods averaged over 10 random seeds with standard deviations shaded in Figure \ref{fig:momdp-results}.
We did not test SARSOP for these domains because the best attainable performance is trivially $1$ by definition.
Ah-Ch + BGN is the only agent that can achieve a perfect success rate in all tasks.
Ab-Cb does significantly worse in a surprising contrast to its classic POMDP performance;
we provide an analysis for this later in Section \ref{sec:bgn_analysis}.
Nevertheless, Ab-Cb still performs better than Ah-Cs, Ah-Cb, and Ah-Ch which all have roughly the same bad performance.
These methods appear to be unable to learn meaningful control policies for these tasks, especially in TopPlate and TwoBumps-2D, and their learning progress is highly unstable compared to Ah-Ch + BGN and Ab-Cb.

\tightsubsection{Robot Evaluation}

We transfer the trained Ah-Ch + BGN policies from the previous section to a real robot.
We use a 2-DoF gripper \citep{schwarm2019floating} mounted on a UR5e robot arm (Figure \ref{fig:robot-setup}), where an impedance controller modulates the compliance of the finger.
Without any fine tuning, all policies have a 100\% success rate on the three tasks.
We describe the behavior of each policy below.
\begin{figure}[t]
    \centering
    \includegraphics[width=0.6\linewidth]{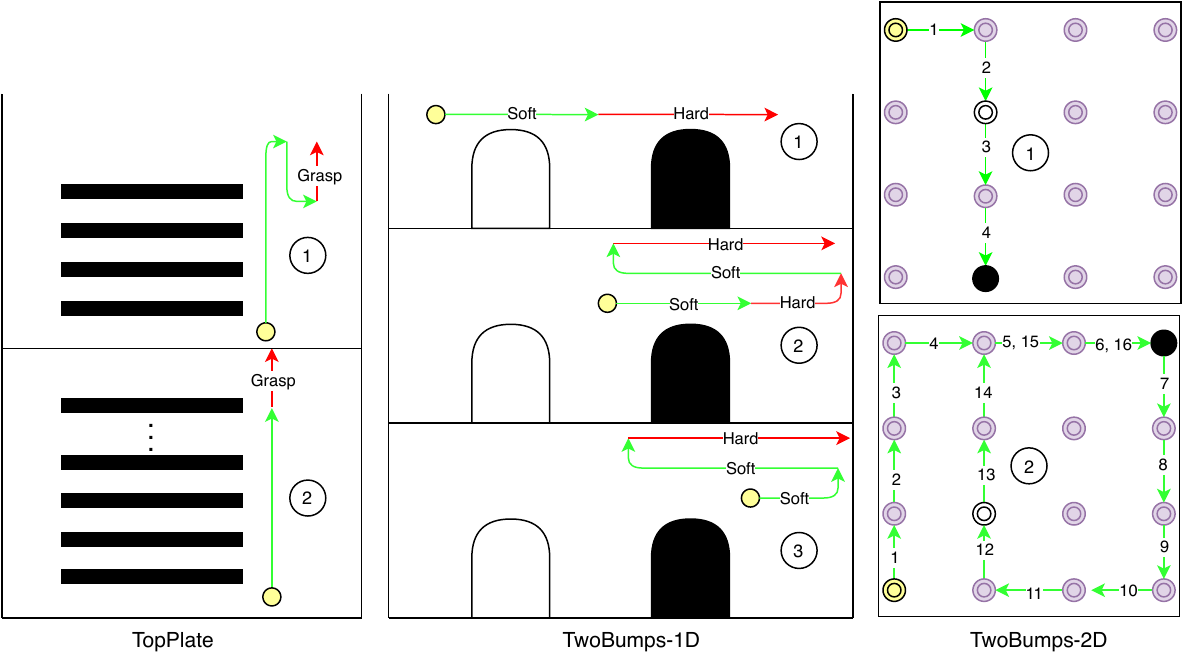}
    \caption{Policies learned by Ah-Ch + BGN in the robotic force-feedback tasks.}
    \label{fig:learned-policies}
\end{figure}

\tightparagraph{TopPlate}
(Figure \ref{fig:learned-policies}, left).
There are two cases.
\numcircled{1}
The number of plates $k$ satisfies $k < 10$, the finger (yellow circle) goes upward until no plate is felt then it moves down one step to grasp the top plate.
\numcircled{2}
When $k=10$, the agent discovers a shortcut:
it grasps the tenth plate immediately after detecting it, indicating that it has learned to count past contact events.

\tightparagraph{TwoBumps-1D}
(Figure \ref{fig:learned-policies}, middle).
The compliant finger (yellow circle) unconditionally moves right at the beginning of the episode.
There are three possible cases.
\numcircled{1}
The finger encounters the first bump, becomes rigid after passing it, and then pushes the second bump to accomplish the goal.
\numcircled{2}
Similar to the first case but the finger reaches the extremity of its motion range before finding the second bump.
The agent realizes that the first bump must therefore be the target bump.
It relaxes and backtracks to the left until it passes the bump again, stiffens, and then returns right to accomplish the goal.
\numcircled{3}
The finger does not initially encounter any bump.
It remains compliant and backtracks to the left until it passes a bump, stiffens, and then returns right to accomplish the goal.

\tightparagraph{TwoBumps-2D}
(Figure \ref{fig:learned-policies}, right).
The soft finger (yellow circle) explores the gridworld efficiently by following a non-intersecting path (green arrows) until it locates both the small and large bumps (white and black circles, respectively).
When the agent encounters the second bump, there are two cases.
\numcircled{1}
The bump is the large one and the agent grasps it immediately (\textit{e.g.}, $t=5$).
\numcircled{2}
The bump is the small one;
the agent traverses the shortest path back to the larger bump and grasps it (\textit{e.g.}, $t=17$).

\subsection{Why Do BGNs Work Well?}
\label{sec:bgn_analysis}

\begin{figure}[t]
    \centering
    \includegraphics[scale=0.4]{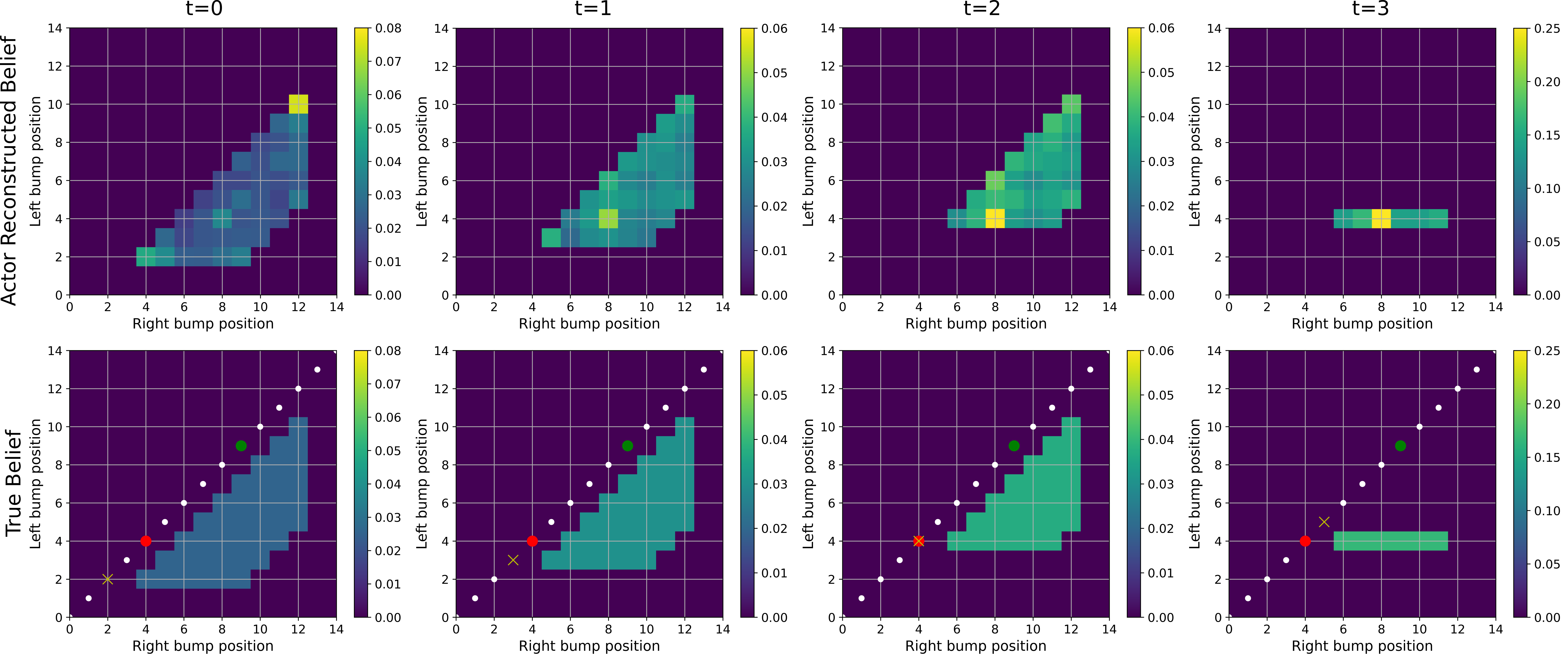}
    \caption{
        Ah-Ch + BGN predicted beliefs (top row) compared with the ground-truth beliefs (bottom row) at various times during execution in TwoBumps-1D.
        The left bump (red), the right bump (green), and the agent (yellow) move diagonally along the permissible locations (white).
    }
    \label{fig:belief-prop}
\end{figure}

Ah-Ch + BGN performs significantly better than the other methods in the force-feedback robot environments.
This disparity cannot be explained by its use of beliefs alone, since both Ah-Cb and Ab-Cb also utilize beliefs.
We identify some the probable mechanisms for this in this section.
To first test how accurately BGN can reconstruct the belief, we compare visualizations of the predicted belief from the trained Ah-Ch + BGN agent against the ground-truth belief in TwoBumps-1D (Figure \ref{fig:belief-prop}).
We see that BGN is able to accurately reproduce the shape of the belief over the two bumps' positions, with only slight variations in magnitude due to approximation errors.
This high-fidelity belief reconstruction indicates that the agent has learned a reasonable representation that can overcome the environment's partial observability.
Given that Ah-Ch + BGN performs significantly better than Ah-Ch, accurate belief prediction appears to be a major factor in determining performance.

Because Ab-Cb also has full access to beliefs during training, it is surprising that it cannot match the performance of Ah-Ch + BGN.
This is particularly unexpected because Ab-Cb and Ah-Ch + BGN performed similarly in the classic POMDPs (Section \ref{subsec:classic}).
A major distinction between these two cases is their relative difference in reward sparsity.
Whereas the classic POMDPs tend to penalize the agent for incorrect actions and guide it towards a good policy, the robot tasks admit positive rewards only upon a successfully completed trial.
Ab-Cb ceases to learn in the absence of rewards---despite having access to the belief---since it has no alternative learning signal.
On the other hand, Ah-Ch + BGN can continue to learn without rewards by reducing the belief-reconstruction loss and improving its feature representation.
When a reinforcement event finally does occur, Ah-Ch + BGN can utilize these better features to learn more effectively.

\begin{figure}[t]
\centering
    \includegraphics[scale=0.6]{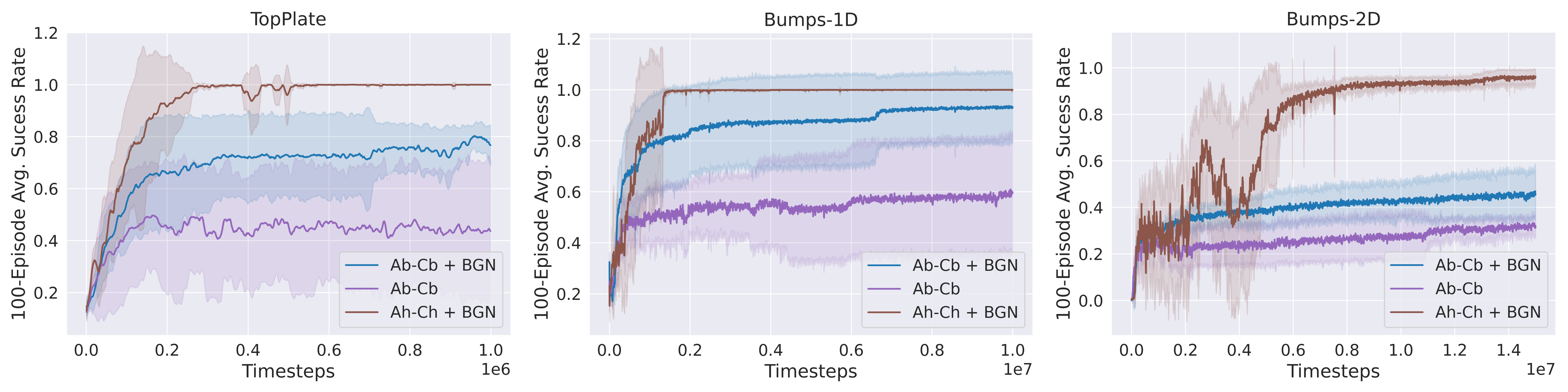}
    \caption{BGN can also improve the performance of an Ab-Cb agent.}
    \label{fig:abcb-bgn}
\end{figure}

If it is true that learning policies directly from beliefs is difficult, then we would expect that adding BGN to Ab-Cb would improve performance.
We repeated the robot experiments with this new method combination, Ab-Cb + BGN, where it is able to recover roughly half of the performance difference between Ab-Cb and Ah-Ch + BGN (Figure \ref{fig:abcb-bgn}).
Still, it is surprising that Ab-Cb + BGN is unable to match the performance of Ah-Ch + BGN.
We hypothesize that the structure of the belief---a large categorical distribution over states---is not necessarily conducive to learning despite being a sufficient statistic for the history.
On the other hand, Ah-Ch + BGN has the advantage of flexibility;
its recurrent neural network can integrate many observations over time and generate a representation that is favorable to the task at hand.
Nevertheless, this final experiment demonstrates that BGN can assist learning regardless of the policy input.

\tightsection{Conclusion}
\label{sec:conclusion}
We introduced Belief-Grounded Networks (BGNs), an end-to-end method for injecting belief information into deep reinforcement learning agents to accelerate their learning under partial observability.
By reconstructing the belief during simulated training, the agent learns to overcome uncertainty in its environment.
Most notably, the BGN does not require beliefs during execution, making the trained policies amenable to physical systems.
When transferred to a real-world robot, these policies masterfully completed all of our proposed manipulation tasks without any fine tuning or other adjustments.

Our work focused on discrete-state environments, but future work could easily extend our method to continuous-state tasks.
For example, the belief update could be approximated by various methods.
One possibility would assume that the belief has some simplifying parametric form, such as a Gaussian mixture model, and then apply an approximate belief update over these parameters.
An alternative would utilize Monte Carlo techniques, such as a particle filter, to maintain an estimate of the belief without requiring a closed-form distribution.
We anticipate there to be many choices of approximations that work well in practice, provided that they can reconstruct the belief with reasonable accuracy.
This flexibility of the BGN is one of its most noteworthy advantages, since the information-theoretic loss is well-defined for any conceivable belief distribution regardless of its precise functional form.



\clearpage
\acknowledgments{We would like to thank Sammie Katt and Andrea Baisero for useful discussions. This material is based upon research supported by, or in part by, the U. S. Office of Naval Research under award number N00014-19-1-2131 and the Army Research Office award W911NF-20-1-0265. It is also supported by NSF grants 1816382, 1830425, 1724257, and 1724191.
}


\small
\bibliography{references}  

\begin{thebibliography}{40}
\providecommand{\natexlab}[1]{#1}
\providecommand{\url}[1]{\texttt{#1}}
\expandafter\ifx\csname urlstyle\endcsname\relax
  \providecommand{\doi}[1]{doi: #1}\else
  \providecommand{\doi}{doi: \begingroup \urlstyle{rm}\Url}\fi

\bibitem[Kurniawati et~al.(2008)Kurniawati, Hsu, and Lee]{Kurniawati_2008}
H.~Kurniawati, D.~Hsu, and W.~S. Lee.
\newblock {SARSOP}: Efficient point-based {POMDP} planning by approximating
  optimally reachable belief spaces.
\newblock In \emph{Robotics: Science and Systems {IV}}. Robotics: Science and
  Systems Foundation, June 2008.

\bibitem[Silver and Veness(2010)]{silver2010monte}
D.~Silver and J.~Veness.
\newblock Monte-carlo planning in large {POMDP}s.
\newblock In \emph{Advances in Neural Information Processing Systems}, pages
  2164--2172, 2010.

\bibitem[Smith and Simmons(2003)]{smith2012heuristic}
T.~Smith and R.~Simmons.
\newblock Heuristic search value iteration for {POMDP}s.
\newblock In \emph{Proceedings of Proc. of UAI 2004}, Banff, Alberta, December
  2003.

\bibitem[Somani et~al.(2013)Somani, Ye, Hsu, and Lee]{somani2013despot}
A.~Somani, N.~Ye, D.~Hsu, and W.~S. Lee.
\newblock {DESPOT}: Online {POMDP} planning with regularization.
\newblock In \emph{Advances in Neural Information Processing Systems}, pages
  1772--1780, 2013.

\bibitem[Hausknecht and Stone(2015)]{hausknecht2015deep}
M.~Hausknecht and P.~Stone.
\newblock Deep recurrent q-learning for partially observable {MDP}s.
\newblock In \emph{2015 AAAI Fall Symposium Series}, 2015.

\bibitem[Pinto et~al.(2018)Pinto, Andrychowicz, Welinder, Zaremba, and
  Abbeel]{pinto2017asymmetric}
L.~Pinto, M.~Andrychowicz, P.~Welinder, W.~Zaremba, and P.~Abbeel.
\newblock Asymmetric actor critic for image-based robot learning.
\newblock In \emph{Robotics: Science and Systems {XIV}}. Robotics: Science and
  Systems Foundation, June 2018.

\bibitem[Kaelbling et~al.(1998)Kaelbling, Littman, and
  Cassandra]{kaelbling1998planning}
L.~P. Kaelbling, M.~L. Littman, and A.~R. Cassandra.
\newblock Planning and acting in partially observable stochastic domains.
\newblock \emph{Artificial Intelligence}, 101\penalty0 (1-2):\penalty0 99--134,
  1998.

\bibitem[Sutton and Barto(2018)]{sutton2018reinforcement}
R.~S. Sutton and A.~G. Barto.
\newblock \emph{Reinforcement Learning: An Introduction}.
\newblock MIT press, 2018.

\bibitem[Dhariwal et~al.(2017)Dhariwal, Hesse, Klimov, Nichol, Plappert,
  Radford, Schulman, Sidor, Wu, and Zhokhov]{dhariwal2017openai}
P.~Dhariwal, C.~Hesse, O.~Klimov, A.~Nichol, M.~Plappert, A.~Radford,
  J.~Schulman, S.~Sidor, Y.~Wu, and P.~Zhokhov.
\newblock {O}pen{AI} baselines, 2017.

\bibitem[Mnih et~al.(2016)Mnih, Badia, Mirza, Graves, Lillicrap, Harley,
  Silver, and Kavukcuoglu]{mnih2016asynchronous}
V.~Mnih, A.~P. Badia, M.~Mirza, A.~Graves, T.~Lillicrap, T.~Harley, D.~Silver,
  and K.~Kavukcuoglu.
\newblock Asynchronous methods for deep reinforcement learning.
\newblock In \emph{International Conference on Machine Learning}, pages
  1928--1937, 2016.

\bibitem[Rumelhart et~al.(1986)Rumelhart, Hinton, and
  Williams]{rumelhart1986learning}
D.~E. Rumelhart, G.~E. Hinton, and R.~J. Williams.
\newblock Learning representations by back-propagating errors.
\newblock \emph{Nature}, 323\penalty0 (6088):\penalty0 533--536, 1986.

\bibitem[Lillicrap et~al.(2015)Lillicrap, Hunt, Pritzel, Heess, Erez, Tassa,
  Silver, and Wierstra]{lillicrap2015continuous}
T.~P. Lillicrap, J.~J. Hunt, A.~Pritzel, N.~Heess, T.~Erez, Y.~Tassa,
  D.~Silver, and D.~Wierstra.
\newblock Continuous control with deep reinforcement learning.
\newblock \emph{arXiv preprint arXiv:1509.02971}, 2015.

\bibitem[Schulman et~al.(2017)Schulman, Wolski, Dhariwal, Radford, and
  Klimov]{schulman2017proximal}
J.~Schulman, F.~Wolski, P.~Dhariwal, A.~Radford, and O.~Klimov.
\newblock Proximal policy optimization algorithms.
\newblock \emph{arXiv preprint arXiv:1707.06347}, 2017.

\bibitem[Mnih et~al.(2015)Mnih, Kavukcuoglu, Silver, Rusu, Veness, Bellemare,
  Graves, Riedmiller, Fidjeland, Ostrovski, et~al.]{mnih2015human}
V.~Mnih, K.~Kavukcuoglu, D.~Silver, A.~A. Rusu, J.~Veness, M.~G. Bellemare,
  A.~Graves, M.~Riedmiller, A.~K. Fidjeland, G.~Ostrovski, et~al.
\newblock Human-level control through deep reinforcement learning.
\newblock \emph{Nature}, 518\penalty0 (7540):\penalty0 529--533, 2015.

\bibitem[Wilson and Hermans(2020)]{wilson2020learning}
M.~Wilson and T.~Hermans.
\newblock Learning to manipulate object collections using grounded state
  representations.
\newblock In \emph{Conference on Robot Learning}, pages 490--502, 2020.

\bibitem[Chen et~al.(2020)Chen, Zhou, Koltun, and
  Kr{\"a}henb{\"u}hl]{chen2020learning}
D.~Chen, B.~Zhou, V.~Koltun, and P.~Kr{\"a}henb{\"u}hl.
\newblock Learning by cheating.
\newblock In \emph{Conference on Robot Learning}, pages 66--75, 2020.

\bibitem[Sygnowski and Michalewski(2016)]{sygnowski2016learning}
J.~Sygnowski and H.~Michalewski.
\newblock Learning from the memory of {A}tari 2600.
\newblock In \emph{Computer Games}, pages 71--85. Springer, 2016.

\bibitem[Pathak et~al.(2017)Pathak, Agrawal, Efros, and
  Darrell]{pathak2017curiosity}
D.~Pathak, P.~Agrawal, A.~A. Efros, and T.~Darrell.
\newblock Curiosity-driven exploration by self-supervised prediction.
\newblock In \emph{Proceedings of the IEEE Conference on Computer Vision and
  Pattern Recognition Workshops}, pages 16--17, 2017.

\bibitem[Jaderberg et~al.(2016)Jaderberg, Mnih, Czarnecki, Schaul, Leibo,
  Silver, and Kavukcuoglu]{jaderberg2016reinforcement}
M.~Jaderberg, V.~Mnih, W.~M. Czarnecki, T.~Schaul, J.~Z. Leibo, D.~Silver, and
  K.~Kavukcuoglu.
\newblock Reinforcement learning with unsupervised auxiliary tasks.
\newblock \emph{arXiv preprint arXiv:1611.05397}, 2016.

\bibitem[Mirowski et~al.(2016)Mirowski, Pascanu, Viola, Soyer, Ballard, Banino,
  Denil, Goroshin, Sifre, Kavukcuoglu, et~al.]{mirowski2016learning}
P.~Mirowski, R.~Pascanu, F.~Viola, H.~Soyer, A.~J. Ballard, A.~Banino,
  M.~Denil, R.~Goroshin, L.~Sifre, K.~Kavukcuoglu, et~al.
\newblock Learning to navigate in complex environments.
\newblock \emph{arXiv preprint arXiv:1611.03673}, 2016.

\bibitem[Shelhamer et~al.(2016)Shelhamer, Mahmoudieh, Argus, and
  Darrell]{shelhamer2016loss}
E.~Shelhamer, P.~Mahmoudieh, M.~Argus, and T.~Darrell.
\newblock Loss is its own reward: Self-supervision for reinforcement learning.
\newblock \emph{arXiv preprint arXiv:1612.07307}, 2016.

\bibitem[Srinivas et~al.(2020)Srinivas, Laskin, and Abbeel]{srinivas2020curl}
A.~Srinivas, M.~Laskin, and P.~Abbeel.
\newblock {CURL}: Contrastive unsupervised representations for reinforcement
  learning.
\newblock \emph{arXiv preprint arXiv:2004.04136}, 2020.

\bibitem[Littman and Sutton(2002)]{littman2002predictive}
M.~L. Littman and R.~S. Sutton.
\newblock Predictive representations of state.
\newblock In \emph{Advances in Neural Information Processing Systems}, pages
  1555--1561, 2002.

\bibitem[Singh et~al.(2004)Singh, James, and Rudary]{singh2004predictive}
S.~Singh, M.~James, and M.~Rudary.
\newblock Predictive state representations: A new theory for modeling dynamical
  systems.
\newblock In \emph{Proceedings of the 20th Conference on Uncertainty in
  Artificial Intelligence}, 2004.

\bibitem[Baisero and Amato(2020)]{baisero2020learning}
A.~Baisero and C.~Amato.
\newblock Learning complementary representations of the past using auxiliary
  tasks in partially observable reinforcement learning.
\newblock In \emph{Proceedings of the 19th International Conference on
  Autonomous Agents and Multi-Agent Systems}, pages 1762--1764, 2020.

\bibitem[Vien and Toussaint(2015)]{vien2015pomdp}
N.~A. Vien and M.~Toussaint.
\newblock {POMDP} manipulation via trajectory optimization.
\newblock In \emph{2015 IEEE/RSJ International Conference on Intelligent Robots
  and Systems (IROS)}, pages 242--249. IEEE, 2015.

\bibitem[{Garg} et~al.(2019){Garg}, {Hsu}, and {Lee}]{garg2019learning}
N.~P. {Garg}, D.~{Hsu}, and W.~S. {Lee}.
\newblock Learning to grasp under uncertainty using {POMDP}s.
\newblock In \emph{2019 International Conference on Robotics and Automation
  (ICRA)}, pages 2751--2757, 2019.

\bibitem[Wirnshofer et~al.(2020)Wirnshofer, Schmitt, Wichert, and
  Burgard]{wirnshofer2020controlling}
F.~Wirnshofer, P.~S. Schmitt, G.~v. Wichert, and W.~Burgard.
\newblock Controlling contact-rich manipulation under partial observability.
\newblock In \emph{Robotics: Science and Systems}, 2020.

\bibitem[Sung et~al.(2017)Sung, Salisbury, and Saxena]{sung2017learning}
J.~Sung, J.~K. Salisbury, and A.~Saxena.
\newblock Learning to represent haptic feedback for partially-observable tasks.
\newblock In \emph{2017 International Conference on Robotics and Automation
  (ICRA)}, pages 2802--2809. IEEE, 2017.

\bibitem[Beltran-Hernandez et~al.(2020)Beltran-Hernandez, Petit,
  Ramirez-Alpizar, Nishi, Kikuchi, Matsubara, and Harada]{beltran2020learning}
C.~C. Beltran-Hernandez, D.~Petit, I.~G. Ramirez-Alpizar, T.~Nishi, S.~Kikuchi,
  T.~Matsubara, and K.~Harada.
\newblock Learning force control for contact-rich manipulation tasks with rigid
  position-controlled robots.
\newblock \emph{arXiv preprint arXiv:2003.00628}, 2020.

\bibitem[Cho et~al.(2014)Cho, Van~Merri{\"e}nboer, Bahdanau, and
  Bengio]{cho2014properties}
K.~Cho, B.~Van~Merri{\"e}nboer, D.~Bahdanau, and Y.~Bengio.
\newblock On the properties of neural machine translation: Encoder-decoder
  approaches.
\newblock \emph{arXiv preprint arXiv:1409.1259}, 2014.

\bibitem[Littman et~al.(1995)Littman, Cassandra, and
  Kaelbling]{littman1995learning}
M.~L. Littman, A.~R. Cassandra, and L.~P. Kaelbling.
\newblock Learning policies for partially observable environments: Scaling up.
\newblock In \emph{Machine Learning Proceedings 1995}, pages 362--370.
  Elsevier, 1995.

\bibitem[Todorov et~al.(2012)Todorov, Erez, and Tassa]{todorov2012mujoco}
E.~Todorov, T.~Erez, and Y.~Tassa.
\newblock {M}u{J}o{C}o: A physics engine for model-based control.
\newblock In \emph{2012 IEEE/RSJ International Conference on Intelligent Robots
  and Systems (IROS)}, pages 5026--5033. IEEE, 2012.

\bibitem[Ong et~al.(2009)Ong, Png, Hsu, and Lee]{ong2009pomdps}
S.~C. Ong, S.~W. Png, D.~Hsu, and W.~S. Lee.
\newblock {POMDP}s for robotic tasks with mixed observability.
\newblock In \emph{Robotics: Science and Systems}, volume~5, page~4, 2009.

\bibitem[Schwarm et~al.(2019)Schwarm, Gravesmill, and
  Whitney]{schwarm2019floating}
E.~Schwarm, K.~M. Gravesmill, and J.~P. Whitney.
\newblock A floating-piston hydrostatic linear actuator and remote-direct-drive
  2-dof gripper.
\newblock In \emph{2019 International Conference on Robotics and Automation
  (ICRA)}, pages 7562--7568. IEEE, 2019.

\bibitem[Daley and Amato(2019)]{daley2019reconciling}
B.~Daley and C.~Amato.
\newblock Reconciling $\lambda$-returns with experience replay.
\newblock In \emph{Advances in Neural Information Processing Systems}, pages
  1133--1142, 2019.

\bibitem[Schulman et~al.(2016)Schulman, Moritz, Levine, Jordan, and
  Abbeel]{schulman2016high}
J.~Schulman, P.~Moritz, S.~Levine, M.~Jordan, and P.~Abbeel.
\newblock High-dimensional continuous control using generalized advantage
  estimation.
\newblock In \emph{Proceedings of the International Conference on Learning
  Representations (ICLR)}, 2016.

\bibitem[Tieleman and Hinton(2012)]{tieleman2012lecture}
T.~Tieleman and G.~Hinton.
\newblock Lecture 6.5-{RMS}prop: Divide the gradient by a running average of
  its recent magnitude.
\newblock \emph{Coursera: Neural networks for machine learning}, 4\penalty0
  (2):\penalty0 26--31, 2012.

\bibitem[Pascanu et~al.(2013)Pascanu, Mikolov, and
  Bengio]{pascanu2013difficulty}
R.~Pascanu, T.~Mikolov, and Y.~Bengio.
\newblock On the difficulty of training recurrent neural networks.
\newblock In \emph{International Conference on Machine Learning}, pages
  1310--1318, 2013.

\bibitem[Brockman et~al.(2016)Brockman, Cheung, Pettersson, Schneider,
  Schulman, Tang, and Zaremba]{brockman2016openai}
G.~Brockman, V.~Cheung, L.~Pettersson, J.~Schneider, J.~Schulman, J.~Tang, and
  W.~Zaremba.
\newblock {O}pen{AI} {G}ym.
\newblock \emph{arXiv preprint arXiv:1606.01540}, 2016.

\end{thebibliography}
\normalsize

\clearpage

\appendix

\section{Training Details}
\label{app:training-details}
Hyperparameters for all agents are listed in Table~\ref{tab:training-params}.
We estimate returns $R_t$ using variable-length $\lambda$-returns that bootstrap from the end of the current trajectory (\textit{e.g.}, see \cite{daley2019reconciling}).
This makes the advantage estimate $A_t = R_t - V(s_t; \phi)$ equivalent to Generalized Advantage Estimation (GAE)~\cite{schulman2016high}.
During training, the entropy loss coefficient $\beta_e$ is held constant in all domains except RockSample, where we found it necessary to decay $\beta_e$ exponentially to obtain good performance.

\def\arraystretch{1.25}
\begin{table}[ht]
    \centering
    \begin{tabular}{l c c}
        \toprule
        \textbf{Hyperparameter} & \textbf{Symbol} & \textbf{Value} \\
        \midrule
        Number of actors & & 16 \\
        Sample length & $T$ & 5 \\
        $\lambda$-return/GAE parameter & $\lambda$ & 0.95 \\
        Critic loss coefficient & $\beta_c$ & 0.5 \\ 
        Entropy loss coefficient & $\beta_e$ & $\begin{cases}
                                                    2 \times 0.1^{\sfrac{t}{3000000}} \text{\quad RockSample} \\
                                                    0.01 \text{\quad\quad\quad\ all other domains}
                                                \end{cases}$ \\
        RMSProp learning rate & & $7 \times 10^{-4}$ \\
        RMSProp decay parameter & & 0.99 \\
        RMSProp denominator constant & & 0.95 \\
        Clipped gradient norm magnitude & & 0.5 \\
        \bottomrule
    \end{tabular}
    \vspace{0.1in}
    \caption{
        A2C hyperparameter values used across all experiments.
        RMSProp is the first-order optimization method proposed in \cite{tieleman2012lecture}.
        During optimization, the $L^2$-norm of each gradient is re-scaled such that it never exceeds the clipped gradient norm magnitude \cite{pascanu2013difficulty}.
    }
    \label{tab:training-params}
\end{table}

\clearpage

\section{Agents}
\label{app:agents}
Tables \ref{tab:architectures-pomdp} and \ref{tab:architectures-robot} summarize the neural network architectures used for the classic POMDPs and force-feedback robots, respectively.
Input concatenation is represented by brackets.
FC($h$) and GRU($h$) denote fully connected layers and Gated Recurrent Units \cite{cho2014properties}, respectively, with $h$ hidden units.
A large brace indicates a branch in the architecture for the belief-reconstruction loss;
this is used only by BGN agents.
The \textit{output-size} is $|\mathcal{A}|$ for actors and $1$ for critics.
The \textit{belief-size} is the dimension of the belief.
The parenthetical term ``(+ softmax)'' indicates that a softmax activation function is used by actors, but a linear activation function is used by critics.

In the classic POMDP domains, states, observations, and actions are represented as abstract indices.
We use Pytorch's embedding layer (essentially, a one-hot encoding followed by FC layer) to convert these indices to fixed-length vectors.
We denote these layers as Embed(input, $h$) where $h$ represents the dimensionality of the output vector.

In the force-feedback robot domains, states and observations have physically meaningful interpretations.
We therefore do not use embedding layers.
Instead, we simply min-max normalize observations, actions, and states into the interval $[0,1]$.

Each agent's architectural choices are uniquely determined by their input types:
\begin{itemize}
    \def\arraystretch{1}
    \item \textbf{Ah-Ch}: Recurrent actor and Recurrent critic. Both use observation-action pairs.
    \item \textbf{Ab-Cb}: Feedforward actor and Feedforward critic. Both use ground-truth beliefs.
    \item \textbf{Ah-Cb}: Recurrent actor and Feedforward critic. The actor uses observation-action pairs while the critic uses ground-truth beliefs.
    \item \textbf{Ah-Cs}: Recurrent actor and Feedforward critic. The actor uses observation-action pairs while the critic uses ground-truth states.
\end{itemize}

\begin{table}[h]
    \def\arraystretch{1.5}
    \centering
    \begin{tabular}{l c c}
        \toprule
        \textbf{Layer} & \textbf{Recurrent} & \textbf{Feedforward} \\
        \midrule
        Input & \def\arraystretch{1.0}
                \begin{tabular}{@{}c@{}}
                    [Embed(Observation, 128),\\
                    Embed(Action, 128)]
                \end{tabular}
                \def\arraystretch{1.5}
              & Belief or Embed(State, 128) \\
        Hidden1 & GRU(256) & FC(256) + tanh \\
        Hidden2 & FC(256) + tanh & FC(256) + tanh \\
        Outputs & \hspace{0.25in} $\begin{cases}
                    \text{FC(output-size) (+ softmax)}\\
                    \text{FC(belief-size) + softmax}
                \end{cases}$
                & \hspace{0.25in} $\begin{cases}
                    \text{FC(output-size) (+ softmax)}\\
                    \text{FC(belief-size) + softmax}
                \end{cases}$ \\ 
        \bottomrule
    \end{tabular}
    \vspace{0.1in}
    \caption{
        Neural network architectures used for the classic POMDP experiments (Section~\ref{subsec:classic}).
    }
    \label{tab:architectures-pomdp}
\end{table}

\begin{table}[h]
    \def\arraystretch{1.5}
    \centering
    \begin{tabular}{l c c}
        \toprule
        \textbf{Layer} & \textbf{Recurrent} & \textbf{Feedforward} \\
        \midrule
        Input & [Observation, Action]
              & State or Belief \\
        Hidden1 & GRU(256) & FC(256) + tanh \\
        Hidden2 & FC(256) + tanh & FC(256) + tanh \\
        Outputs & \hspace{0.25in} $\begin{cases}
                    \text{FC(output-size) (+ softmax)}\\
                    \text{FC(belief-size) + softmax}
                \end{cases}$
                & \hspace{0.25in} $\begin{cases}
                    \text{FC(output-size) (+ softmax)}\\
                    \text{FC(belief-size) + softmax}
                \end{cases}$ \\ 
        \bottomrule
    \end{tabular}
    \vspace{0.1in}
    \caption{
        Neural network architectures used for the force-feedback robot experiments (Section~\ref{subsec:tactile}).
    }
    \label{tab:architectures-robot}
\end{table}

\clearpage

\section{Domain Details}
\begin{figure}[h]
    \centering
    \includegraphics[scale=0.55]{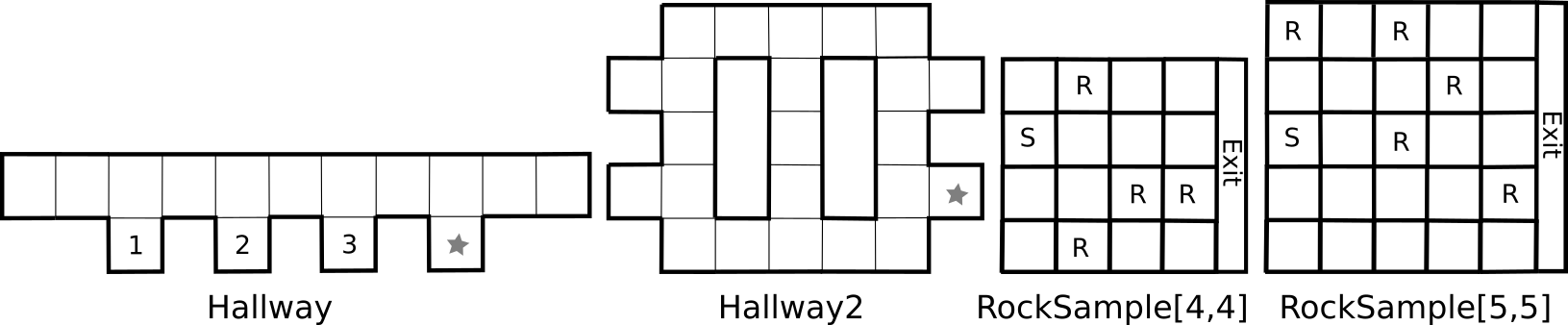}
    \caption{
        Maps for the classic POMDP domains.
    }
    \label{fig:pomdp-domains}
\end{figure}

\tightsubsection{Classic POMDP domains}
\label{app:pomdp-details}
\paragraph{Hallway and Hallway-2}
The maps for Hallway and Hallway-2 are shown in Figure \ref{fig:pomdp-domains} with the target cell denoted by $\star$.
In Hallway, there are 3 landmarks that will be visible to the agent when it faces south at three particular cells.
In both domains, there are 5 actions available:
\{\textit{stay}, \textit{move forward}, \textit{turn right}, \textit{turn left}, \textit{turn around}\}.
The agent gets a reward of $+1$ when it reaches the target.

\tightparagraph{RockSample[4,4] and RockSample[5,5]}
The rover starts at $S$ and rocks are located in cells marked by $R$ as shown in Figure \ref{fig:pomdp-domains}.
The rover can select from $k+5$ actions:
\{\textit{go north}, \textit{go south}, \textit{go east}, \textit{go west}, \textit{sample}, \textit{$\text{check}_0$}, $\ldots$, \textit{$\text{check}_{k-1}$}\}.
The \textit{sample} action checks the value of the rock at the rover's current location.
The action $\text{check}_i$ examines the quality of the rock at location $i$ with a probabilistic accuracy that decays exponentially as the distance $d$ increases.
The agent gets a reward of $+10$ if it samples a good rock, and then the rock becomes bad.
Sampling a bad rock will give a penalty of $-10$, while moving to the exit area yields a reward of $+10$.
Other moves do not provide any reward.

\tightparagraph{.POMDP files}
We obtain the .POMDP files that define Hallway domains from \url{https://cs.brown.edu/research/ai/pomdp/examples/index.html}.
For details regarding the file format, we refer readers to \url{http://www.pomdp.org/code/pomdp-file-spec.html}.
We use the script in \url{https://github.com/trey0/zmdp} to generate the files for RockSample domains.
These files initially define the domains as continuing tasks, so we add an \texttt{reset} keyword to convert them into episodic tasks following the format in the \texttt{gym-pomdp} package at \url{https://github.com/abaisero/gym-pomdps}.
This package also converts the domains into a Gym-compatible \cite{brockman2016openai} interface.

\tightparagraph{Parameters}
Table \ref{tab:pomdp-domain-table} contains the parameters of these domains:
the sizes of the state, action, and observation spaces; the discount values; and maximum episode lengths.
During training, we terminate any episode that reaches the maximum episode length.

\tightsubsection{Force-feedback robot domains}
\label{app:momdp-details}
Table \ref{tab:momdp-domain-table} contains the parameters of the force-feedback robot domains:
set sizes, the discount values, and maximum episode lengths.
These domains are classified as MOMDPs (see Appendix \ref{app:momdp}).
During training, we terminate any episode that reaches the maximum episode length.

\begin{table}[h]
    \def\arraystretch{1.0}
    \centering
    \begin{tabular}{c c c c c c c}
        \toprule
        Domain & $|\mathcal{S}|$ & $|\mathcal{A}|$ & $|\Omega|$ & $\gamma$ & Max episode length \\
        \midrule
        Hallway & 57 & 5 & 21 & 0.95 & 100 \\
        Hallway-2 & 89 & 5 & 17 & 0.95 & 100 \\        
        RockSample[4,4] & 257 & 9 & 2 & 0.95 & 100 \\
        RockSample[5,5] & 807 & 10 & 2 & 0.95 & 100 \\
        \bottomrule
    \end{tabular}
    \vspace{0.1in}
    \caption{Classic POMDP domains properties.}
    \label{tab:pomdp-domain-table}
    \begin{tabular}{c c c c c c c c}
        \toprule
        Robot domain & $|\mathcal{X}|$ & $|\mathcal{Y}|$& $|\mathcal{A}|$ & $|\Omega|$ & $\gamma$ & Max episode length \\
        \midrule
        TopPlate & 21 & 8 & 3 & 21 & 0.99 & 50 \\
        TwoBumps-1D & 45 & 225 & 4 & 25 & 0.99 & 100 \\        
        TwoBumps-2D & 48 & 256 & 5 & 48 & 0.99 & 100 \\
        \bottomrule
    \end{tabular}
    
    \vspace{0.1in}
    \caption{Force-feedback robot domains properties.}
    \label{tab:momdp-domain-table}
\end{table}

\clearpage

\section{Raw and Normalized Returns}
\label{app:raw-score}
The average raw returns over 10 random seeds are shown in Table \ref{tab:results-raw}.
The error indicated is one standard deviation around the mean.
We use the code at \url{https://github.com/AdaCompNUS/sarsop} to compute the performance of SARSOP, averaged over $1000$ episodes.

Table \ref{tab:results-normalized} contains the normalized returns computed from the raw returns above with $1$ for SARSOP and $0$ for the random agent:
\textit{Normalized Return} = (\textit{Raw Return} - \textit{Random}) / (\textit{SARSOP} - \textit{Random}).

\begin{table}[h]
    \centering
    \resizebox{\textwidth}{!}
    {
    \begin{tabular}{l c c c c c c c}
        \toprule
        Domain & Ah-Ch & Ah-Ch+BGN & Ah-Cs & Ab-Cb & Ah-Cb & SARSOP & Random\\
        \midrule
        Hallway & 0.50$\pm$0.02 & \textbf{0.54$\pm$0.01} & 0.46$\pm$0.08& 0.53$\pm$0.04 & 0.48$\pm$0.03 & 0.53$\pm$0.05 & 0.05$\pm$0.15\\
        Hallway-2 & 0.1$\pm$0.02 & 0.35$\pm$0.01 & 0.26$\pm$0.02 & \textbf{0.36$\pm$0.01} & 0.29$\pm$0.02 & 0.35$\pm$0.04 & 0.03$\pm$0.11\\        
        RockSample[4,4] & 7.79$\pm$1.12 & 14.14$\pm$0.21 & 6.73$\pm$0.15 & 13.94$\pm$0.33 & 6.1$\pm$2.04& \textbf{17.75$\pm$0.12} & -62.0$\pm$38.2\\
        RockSample[5,5] & 6.23$\pm$0.35 & 14.71$\pm$0.5 & 6.52$\pm$0.09 & 15.22$\pm$0.52 & 6.53$\pm$0.08 & \textbf{19.2$\pm$0.07} & -61.0$\pm$36.12\\
        \bottomrule
    \end{tabular}
    }
    \vspace{0.1in}
    \caption{Raw returns of all agents in the classic POMDP domains, averaged over 10 runs.}
    \label{tab:results-raw}
\end{table}

\begin{table}[h]
    \centering
    \begin{tabular}{l c c c c c}
        \toprule
        Domain & Ah-Ch & Ah-Ch+BGN & Ah-Cs & Ab-Cb & Ah-Cb\\
        \midrule
        Hallway & 0.94 $\pm$ 0.04 & \textbf{1.02 $\pm$ 0.02} & 0.85 $\pm$ 0.17 & 1.0 $\pm$ 0.08 & 0.90 $\pm$ 0.06\\
        Hallway-2 & 0.22 $\pm$ 0.06 & 1.0 $\pm$ 0.03 & 0.72 $\pm$ 0.06 & \textbf{1.03 $\pm$ 0.03} & 0.81 $\pm$ 0.06 \\        
        RockSample[4,4] & 0.88 $\pm$ 0.01 & \textbf{0.95 $\pm$ 0.00} & 0.86 $\pm$ 0.01 & \textbf{0.95 $\pm$ 0.00} & 0.85 $\pm$ 0.03\\
        RockSample[5,5] & 0.84 $\pm$0.00 & 0.94 $\pm$ 0.00 & 0.84 $\pm$ 0.00 & \textbf{0.95 $\pm$ 0.00} & 0.84 $\pm$ 0.00 \\
        \bottomrule
    \end{tabular}
    \vspace{0.1in}
    \caption{
        Normalized returns of agents with 1.0 for SARSOP and 0.0 for the random agent, averaged over 10 runs.
    }
    \label{tab:results-normalized}
\end{table}

\section{Mixed Observability Markov Decision Process (MOMDP)}
\label{app:momdp}
The MOMDP is a generalization of the POMDP where each state $s = (x,y)$ is decomposed into fully and partially observable components $x$ and $y$, respectively.
Thus, the state space is factored as
$\mathcal{S} = \mathcal{X} \times \mathcal{Y}$
with the corresponding transition models
$\mathcal{T}_\mathcal{X}(x,y,a,x') = P(x' \mid x,y,a)$
and
$\mathcal{T}_\mathcal{Y}(x,y,a,x',y') = P(y' \mid x,y,a,x')$.
The observation model becomes
${\mathcal{O}(x',y',a,o)} = {P(o \mid x',y',a)}$.
The MOMDP is therefore fully specified by the tuple
($\mathcal{X}, \mathcal{Y}, \mathcal{A}, \mathcal{T}_\mathcal{X}, \mathcal{T}_\mathcal{Y}, \mathcal{R}, \Omega, \mathcal{O}$).
Upon taking an action $a \in \mathcal{A}$ and observing a new state $s'=(x',y')$ and a new observation $o \in \Omega$,
the belief $b = (x, b_\mathcal{Y})$ is updated according to the following rule:
\begin{equation*}
    b'_\mathcal{Y}(y') \gets
        \frac{\mathcal{O}(x',y',a,o) \sum_{y\in \mathcal{Y}} \mathcal{T}_\mathcal{X}(x,y,a,x') \mathcal{T}_\mathcal{Y}(x,y,a,x',y') b_\mathcal{Y}(y)}
        {P(o \mid a, b)}
\end{equation*}
Hence, the new belief is given by $b' = (x', b'_\mathcal{Y})$.
The principal advantage of the MOMDP framework is that this update becomes significantly more efficient than Equation~(\ref{eq:belief_update}) when $|\mathcal{Y}| \ll |\mathcal{S}|$.

\clearpage

\end{document}